\pdfoutput=1

\documentclass[letterpaper, 10 pt, conference]{ieeeconf}
\usepackage[T1]{fontenc}
\usepackage{cite}
\usepackage{amssymb,amsfonts}
\usepackage{algorithmic}
\usepackage{graphicx}
\usepackage{textcomp}
\usepackage{changes}
\usepackage{subcaption}
\usepackage{xcolor}
\usepackage{color, soul}
\usepackage{amsmath}
\usepackage{booktabs}
\usepackage{multirow}
\usepackage{xstring}
\usepackage{hhline}
\usepackage{siunitx}
\usepackage{physics}
\usepackage{tikz}
\def\checkmark{\tikz\fill[scale=0.4](0,.35) -- (.25,0) -- (1,.7) -- (.25,.15) -- cycle;} 

\IEEEoverridecommandlockouts                              

\usepackage{cite}
\usepackage{caption}
\newcommand{\rom}[1]{\uppercase\expandafter{\romannumeral #1\relax}}

\title{\LARGE \bf
MSDPN: Monocular Depth Prediction with Partial Laser Observation using Multi-stage Neural Networks}

\author{Hyungtae Lim$^{1}$ , Hyeonjae Gil$^{2}$, Hyun Myung$^{3}$, \textit{Senior Member, IEEE}
\thanks{$^{1}$Hyungtae Lim and $^{3}$Hyun Myung are with School of Electrical Engineering, KI-AI, KI-R at KAIST (Korea Advanced Institute of Science and Technology) and $^{2}$Hyeonjae Gil is a undergraduate intern of the laboratory at KAIST, Daejeon, 34141, South Korea. {\tt\small \{shapelim, now9728, jungmokoo, hmyung\}@kaist.ac.kr} \hfill \break 
\indent This work was supported in part by the Industrial Convergence Core Technology Development Program (Development of Robot Intelligence Technology for Mobility with Learning Capability Toward Robust and Seamless Indoor and Outdoor Autonomous Navigation) funded by the Ministry of Trade, Industry and Energy (MOTIE), South Korea, under Grant 10063172 and in part by the Industry Core Technology Development Project (Development of Artificial Intelligence Robot Autonomous Navigation Technology for Agile Movement in Crowded Space) funded by the MOTIE, South Korea, under Grant 20005062.}%
}

\begin{document}

\captionsetup[figure]{labelformat={default},labelsep=period,name={fig.}}

\maketitle
\thispagestyle{empty}
\pagestyle{empty}

\begin{abstract}

In this study, \textcolor{black}{a} deep\textcolor{black}{-}learning-based multi-stage network architecture \textcolor{black}{called} Multi-\textcolor{black}{S}tage Depth Prediction Network (MSDPN) is proposed to predict \textcolor{black}{a} dense depth map using a 2D LiDAR and a monocular camera. 
Our proposed network consists of a multi-stage encoder-decoder architecture and Cross Stage Feature Aggregation (CSFA). The proposed multi-stage encoder-decoder architecture alleviates the partial \textcolor{black}{observation} problem caused by the characteristics of a 2D LiDAR, and CSFA prevents the multi-stage network from diluting the features and allows the network to learn the inter-spatial relationship \textcolor{black}{between features better. Previous works use sub-sampled data from the ground truth as an input rather than actual 2D LiDAR data. In contrast, our approach \textcolor{black}{trains} the model and conducts experiments \textcolor{black}{with} a physically-collected 2D LiDAR dataset.}
\textcolor{black}{To this end}, we acquired our own \textcolor{black}{dataset called} KAIST RGBD-scan dataset and validated the effectiveness and the robustness of MSDPN  \textcolor{black}{under} realistic conditions.
As verified experimentally, our network yields promising performance against state-of-the-art methods. 
Additionally, we analyzed the performance of \textcolor{black}{different input methods} and confirmed that the reference  \textcolor{black}{depth map} is robust in \textcolor{black}{untrained} scenarios. 
\end{abstract}

\section{INTRODUCTION}

Three-dimensional perception plays a key role in both robotics and computer vision fields, such as Simultaneous Localization and Mapping (SLAM), autonomous driving, object recognition, and scene understanding \cite{durrant2006simultaneous, bailey2006simultaneous}. Many research works have utilized a 3D LiDAR sensor \cite{chen2019suma++, li2016vehicle} for its high accuracy and long maximum measurement. However, since 3D LiDARs are cost-prohibitive, other existing several depth sensors, e.g. stereo cameras, structured-light-based depth sensors, are also utilized in many tasks  \textcolor{black}{as} inexpensive alternative\textcolor{black}{s} \cite{khattak2018vision, eitel2015multimodal,lee2018depth, kim2018rgb}. 
Among them, 2D LiDAR sensors are often exploited on mobile robot platforms, since \textcolor{black}{they can} not only perceive surroundings with a wide field of view \textcolor{black}{but also} have high accuracy thanks to their laser-based measurement principle \cite{kim2019gp, cherubini2014autonomous}.

In the meanwhile, numerous researchers \textcolor{black}{in computer vision} have conducted studies to use \textcolor{black}{a monocular camera} for depth estimation  \textcolor{black}{due to} its low cost, lightweight, and energy-efficiency \cite{saxena20083}. Particularly, with the rapid development of research in deep learning, many learning-based studies have been proposed for depth prediction using monocular cameras \cite{garg2016unsupervised, godard2017unsupervised, eigen2014depth}, producing significant results. 
However, these approaches have \textcolor{black}{a} \textit{global scale ambiguity} issue: depth estimation from the network might be unreliable for robotics applications since monocular images cannot provide global\textcolor{black}{-scale measurements} directly \cite{eigen2014depth,liao2017parse}. 

 \begin{figure}[h]
    \centering
    \begin{subfigure}[b]{.95\linewidth}
        \includegraphics[width=\linewidth]{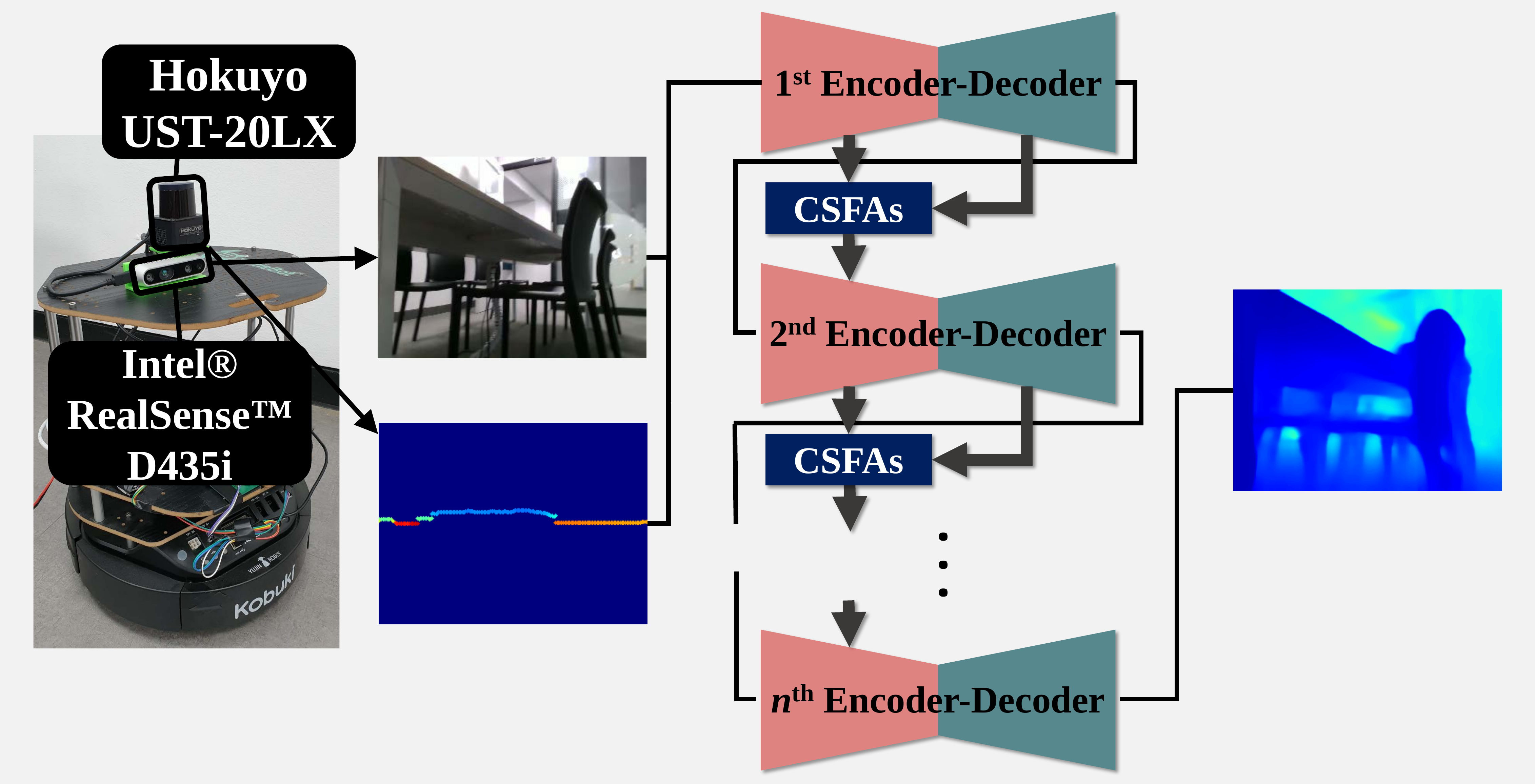}
    \end{subfigure}
    \caption{An \textcolor{black}{o}verview of the proposed learning-based framework.}
    \label{fig:overview_w_comparison}
    
\end{figure}

To alleviate this issue, some researchers have \textcolor{black}{conducted} depth completion tasks, a.k.a sparse-to-dense depth conversion \cite{lee2019depth, mal2018sparse, cheng2018depth}. Using sparse depth on the image plane, they not only mitigate scale ambiguity \textcolor{black}{but also show significant improvement} in depth prediction. Ma and Karaman \cite{mal2018sparse} conduct\textcolor{black}{ed} thorough analysis \textcolor{black}{on} the performance of the deep neural network with different number of sparse depth and suggest\textcolor{black}{ed} many applications. Lee \textit{et al.} \cite{lee2019depth} and Zhang and Funkhouser \cite{zhang2018activestereonet} suggest\textcolor{black}{ed} novel deep feature-guided methods. Cheng \textit{et al.} \cite{cheng2018depth} suggest\textcolor{black}{ed} efficient spatial feature propagation module and show\textcolor{black}{ed} promising performance improvement.

By extension, Liao \textit{et al.} \cite{liao2017parse} and Yin \textit{et al.} \cite{yin2019fusionmapping} suggest\textcolor{black}{ed} novel methods for predicting depth using an image and 2D LiDAR scans to mitigate the limitation of 2D LiDAR scans: \textcolor{black}{extremely} limited vertical field of view. Unlike sparse-to-dense works, the authors reported that the network can hardly learn useful information from the projected depth image generated by 2D laser scans due to the extremely partially sparse distribution, i.e. most range measurements are concentrated horizontally on the image plane. Besides, Liao \textit{et al.} \textcolor{black}{argued} that coexistence of zero-padded values\textcolor{black}{,} which are kind of boolean values to denote whether the laser range data \textcolor{black}{are} projected or not, and valid depth values might confuse the network.

For these reasons, Liao \textit{et al.} \cite{liao2017parse} introduce\textcolor{black}{d} a novel input-level depth information propagation method, called \textit{reference depth map}, which is generated by extending projected 2D observation\textcolor{black}{s in the image plane} along the gravity direction. Through the reference depth map and their proposed networks, they resolve\textcolor{black}{d} the partial observation issues and show\textcolor{black}{ed} prominent performance improvement.

In this paper, a multi-stage network with Cross Stage Feature Aggregation (CSFA) module called Multi-\textcolor{black}{S}tage Depth Prediction Network (MSDPN) is proposed \textcolor{black}{as shown in Fig.~\ref{fig:overview_w_comparison}. }
To the best of our knowledge, it is the first approach that applies multi-stage architecture to mitigate the partial observation problem in realistic environments \textcolor{black}{and, in contrast with previous works \cite{liao2017parse, yin2019fusionmapping}, \textcolor{black}{it utilizes a} physically-collected 2D LiDAR dataset.}
Unlike previous single-stage networks which \textcolor{black}{downsample and upsample} the feature only \textcolor{black}{once}, the MSDPN employs the network-level depth information propagation method to propagate \textcolor{black}{partially} distributed spatial information in the feature to upward and downward regions through repeated upsampling and downsampling.

The contribution of this paper is threefold:
\begin{itemize}
	\item The multi-stage network on which CSFAs are attached is proposed to allow the neural network to prevent saturation and to learn \textcolor{black}{the inter-spatial relationship} of \textcolor{black}{the features} so that the MSDPN yields the better and acceptable depth map compared to previous approaches \textcolor{black}{with fewer number of parameters.}
	
	\item Unlike previous works which \textcolor{black}{use sub-sampled data as if they were obtained from an actual 2D LiDAR}, this paper has trained the model and conducted research using a physically collected dataset from \textcolor{black}{a} 2D LiDAR. Accordingly, we established \textcolor{black}{an} indoor dataset using a 2D LiDAR and a RGB-D camera, \textcolor{black}{called} KAIST RGBD-scan dataset.

	\item \textcolor{black}{We also verify how each component of the MSDPN affects \textcolor{black}{the} performance in \textcolor{black}{the} ablation study. Next, \textcolor{black}{the performance comparison} with \textcolor{black}{different} number of laser scans \textcolor{black}{and different} types of input method are investigated. \textcolor{black}{Finally,} we suggest \textcolor{black}{an} appropriate input method for each 2D LiDAR sensor in realistic applications.}

\end{itemize}

The rest of the paper is organized as follows: Section II introduces the method for projecting laser scans on the image plane which is utilized in acquiring KAIST RGBD-scan dataset. Section III presents our proposed multi-stage neural network in detail. Section IV describes the experiments, and Section V examines the experimental results. Finally, Section \rom{6} summarizes our contributions and describes future works.





\section{Sensor Systems}

\begin{figure}[h]
	\centering
	\begin{subfigure}[b]{0.22\textwidth}
		\includegraphics[width=0.9\textwidth]{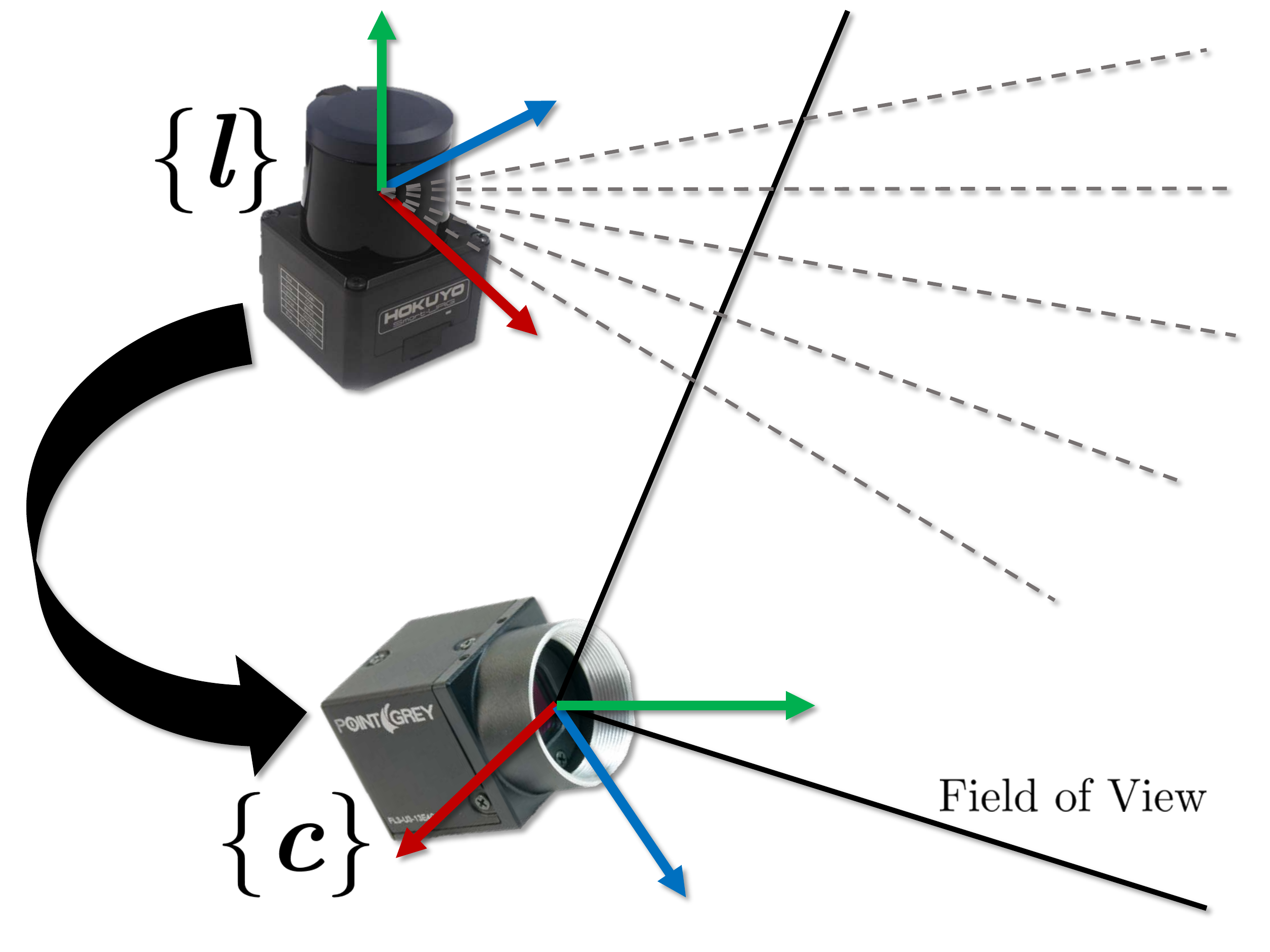}
		\caption{}
		\label{fig:corner_case}
	\end{subfigure}
	\begin{subfigure}[b]{0.25\textwidth}
		\includegraphics[width=0.9\textwidth]{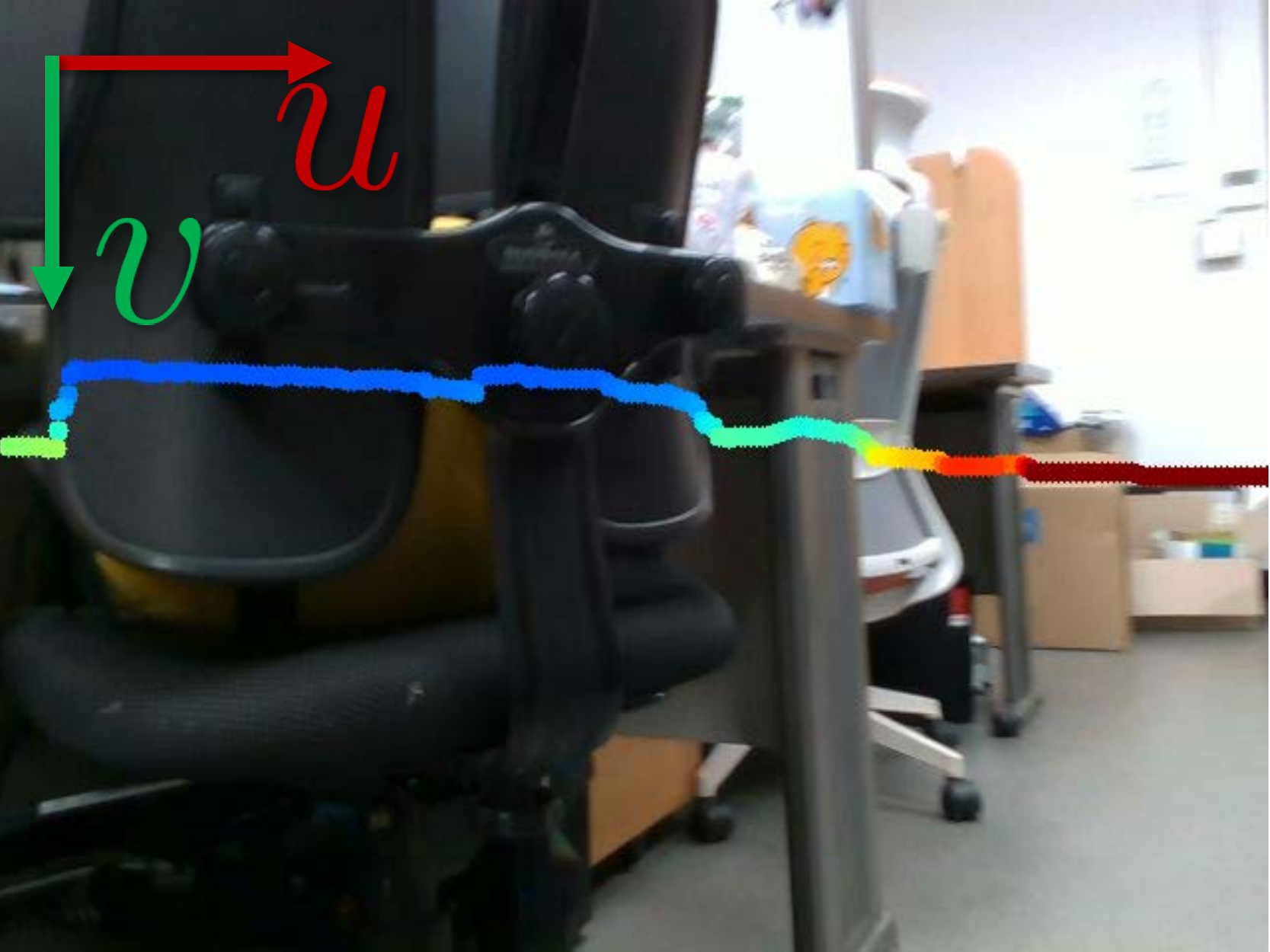}
		\caption{}
		\label{fig:hist_min_idxes}
	\end{subfigure}
	
	\caption{(a) An \textcolor{black}{o}verview of the system. (b) A visualized example \textcolor{black}{where the} projected depth (\texttt{proj-d}) \textcolor{black}{is} overlapped on the image.}\label{fig:sensor_chracteristics}
\end{figure}

In this section, the characteristics of the sensor system which consists of a 2D LiDAR and a monocular camera will be examined, as illustrated in Fig. \ref{fig:sensor_chracteristics}.
Specifically, the method for projecting 2D LiDAR scans onto the image plane of \textcolor{black}{the} RGB camera is explained, which is utilized in acquiring KAIST RGBD-scan dataset (See Section \rom{4}.\textit{A}), and the distribution of the \textcolor{black}{laser scan hits} on the image plane \textcolor{black}{is} analyzed. For simplicity, we use \texttt{proj-d} for the projected depth image, \texttt{RGB} for the RGB image, and \texttt{ref-d} for the reference depth map \cite{liao2017parse}.

 To project laser scans on the image plane, the extrinsic and intrinsic parameters should be estimated. Let $\textbf{p}^{l}=(x^{l}, y^{l}, z^{l})^{T}$ and $\textbf{p}^{c}=(x^{c}, y^{c}, z^{c})^{T}$ be each point in \textcolor{black}{the} 2D LiDAR coordinate system and in \textcolor{black}{the} camera coordinate \textcolor{black}{system}, respectively. Let \textcolor{black}{us denote} a rigid body transformation \textcolor{black}{as} $\mathcal{T} \in SE(3)$ \textcolor{black}{that consists} of corresponding \textcolor{black}{rotation} $\textbf{R}\in SO(3)$ and \textcolor{black}{translation} $\textbf{t} \in \mathbb{R}^{3}$. \textcolor{black}{T}he relationship between $\textbf{p}^{c}$ and $\textbf{p}^{l}$ can be formulated as:
  
\begin{equation}
\textbf{p}^{c} = \textbf{R}\textbf{p}^{l} + \textbf{t}.
\end{equation}
 
Next, according to a pinhole camera model, $\textbf{p}^{c}$ is projected onto the $\textbf{p}^{pixel}=(u, v)^{T}$ by triangulation, where $\textbf{p}^{pixel}$ denotes \textcolor{black}{the} pixel coordinate in the image plane. The corresponding equations are as follows:

\begin{equation}
s
\begin{bmatrix}
u\\
v\\
1
\end{bmatrix} = \textbf{K}\textbf{p}^{c} = 
\begin{bmatrix}
f_{x} & \alpha & c_{x}\\
0 & f_{y} & c_{y} \\
0 & 0 & 1 \\
\end{bmatrix}
\begin{bmatrix}
x^{c}\\
y^{c}\\
z^{c}
\end{bmatrix}
\end{equation}
where $\textbf{K}$ is the camera\textcolor{black}{'s} intrinsic matrix consisting of focal lengths $f_{x}$, $f_{y}$, \textcolor{black}{a} principal point $(c_{x}, c_{y})^{T}$, and \textcolor{black}{a} skew coefficient $\alpha$. $s$ denote\textcolor{black}{s the} scale factor of the image plane. It \textcolor{black}{should be} noted that the scans located beyond \textcolor{black}{the} field of view (FOV) of the camera are filtered out.
 
In short, when value of $y_{c}$ is not large, i.e. when a 2D LiDAR sensor is deployed close to a camera on the y-axis of \textcolor{black}{the} camera coordinate, $v$ value is close to $c_{y}$ in the majority of points. For example, in KAIST RGBD-scan dataset, \textcolor{black}{the} mean and \textcolor{black}{the} standard deviation of \textcolor{black}{the} minimum $v$ positions on the image plane \textcolor{black}{are} 101.2 and 8.8, \textcolor{black}{respectively} when the input size is resized to $304\times228$. Furthermore, the top 90\% (10th percentile) exists between 89 and 117 on the $v$-axis. 

\begin{figure}[h]
    \centering
    \begin{subfigure}[b]{.95\linewidth}
        \includegraphics[width=\linewidth]{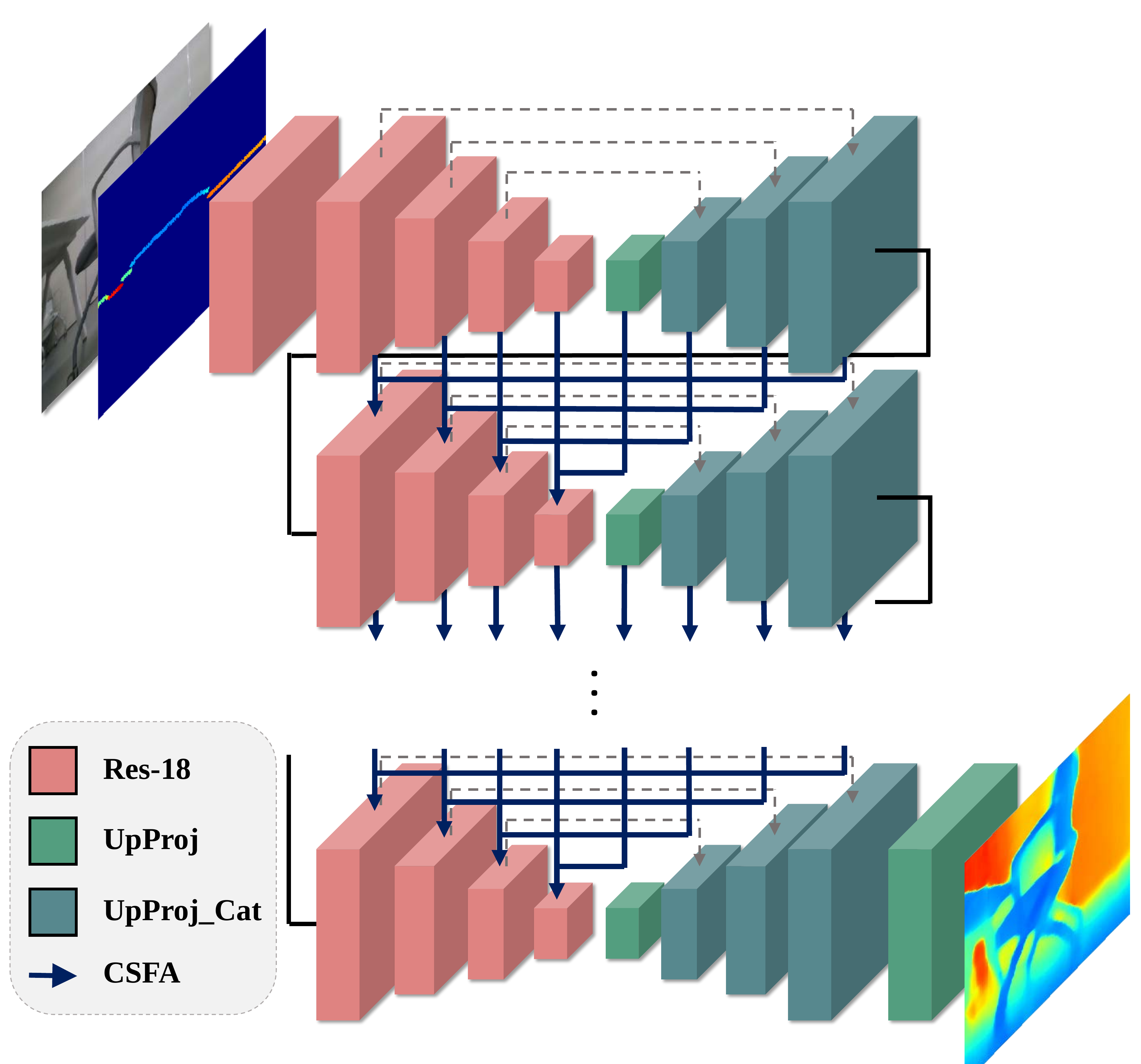}
    \end{subfigure}
    \caption{Our Multi-Stage Depth Prediction Network (MSDPN). The networks is composed of a) multi-stage U-shaped architecture whose backbone is ResNet-18 and b) Cross Stage Feature Aggregation (zoomed-in Figure \ref{fig:csfa}) (best viewed in color).}
    \label{fig:our_network}
    \vspace{-0.5cm}
\end{figure}

\section{Network Architecture}

In this section, our proposed Multi-Stage Depth Prediction Network (MSDPN) will be explained. Specifically, we introduce multi-stage encoder-decoder\textcolor{black}{-}based network and Cross Stage Feature Aggregation (CSFA) to \textcolor{black}{propagate} feature\textcolor{black}{s} to the next stage for more well-described feature\textcolor{black}{s}. The encoding part is based on ResNet-18 \cite{he2016deep}, whose last average pooling and linear transformation layer are neglected, and \texttt{UpProj} \cite{laina2016deeper} and \texttt{UpProj\_Cat} \cite{cheng2018depth} are used in the decoder part.
\textcolor{black}{The single module structure itself does not have novelty, but it is the first time to apply it in a multi-stage encoder-decoder network with CSFA to resolve the partial observation issues effectively.} Besides, through the ablation study, it is shown that the modules \textcolor{black}{not only require fewer parameters} but also show better performance compared to single-stage architectures (See Section \rom{5}.\textit{B} and Table \ref{table:kaist_comparison})

\subsection{Multi-stage Architecture} 

As illustrated in Fig. \ref{fig:our_network}, our multi-stage network consists of $N$ encoder-decoder\textcolor{black}{-}based neural networks. Compared to \textcolor{black}{a} single-stage network, which extract\textcolor{black}{s} representative features in the down \textcolor{black}{sampling} process and recover\textcolor{black}{s} \textcolor{black}{the} lost information in the upsampling procedure \textcolor{black}{only once}, \textcolor{black}{a} multi-stage architecture is able to estimate the final depth with the refined feature. The aforementioned partial observation problem, \textcolor{black}{where} 2D laser scans are mostly located \textcolor{black}{horizontally on the center} of the image, \textcolor{black}{causes spatial information imbalance of the feature}. In other words, only the features in the middle contain  \textcolor{black}{actual} range information \textcolor{black}{directly obtained by the laser scans}, whereas the other features are extracted by \texttt{RGB}. 

\textcolor{black}{For example, many single-stage studies utilize \textcolor{black}{the} ResNet-50\cite{he2016deep} as the backbone \cite{mal2018sparse, cheng2018depth, lee2019depth}. When the input is fed into the ResNet-50-based architecture \textcolor{black}{for} KAIST RGBD-scan dataset, the direct depth cue\textcolor{black}{s} completely propagate \textcolor{black}{from the middle} to \textcolor{black}{the} upward and downward regions of feature after passing through \texttt{conv4\_x}. That is to say, the spatial information is completely propagated when the feature \textcolor{black}{reaches the} final layer of the encoder part.}

Unlike single-stage architecture\textcolor{black}{s}, our multi-stage networks consist of a number of encoder-decoder architecture\textcolor{black}{s}, so the depth information within feature becomes less uneven due to the repeated upsampling and downsampling. As a consequence, this repetition alleviate\textcolor{black}{s} the partial observation issue so that relatively well-balanced feature compared to that of single-stage networks could help the network predict refined depth.

\subsection{Cross Stage Feature Aggregation}

\begin{figure}[h]
    \centering
    \begin{subfigure}[b]{.95\linewidth}
        \includegraphics[width=\linewidth]{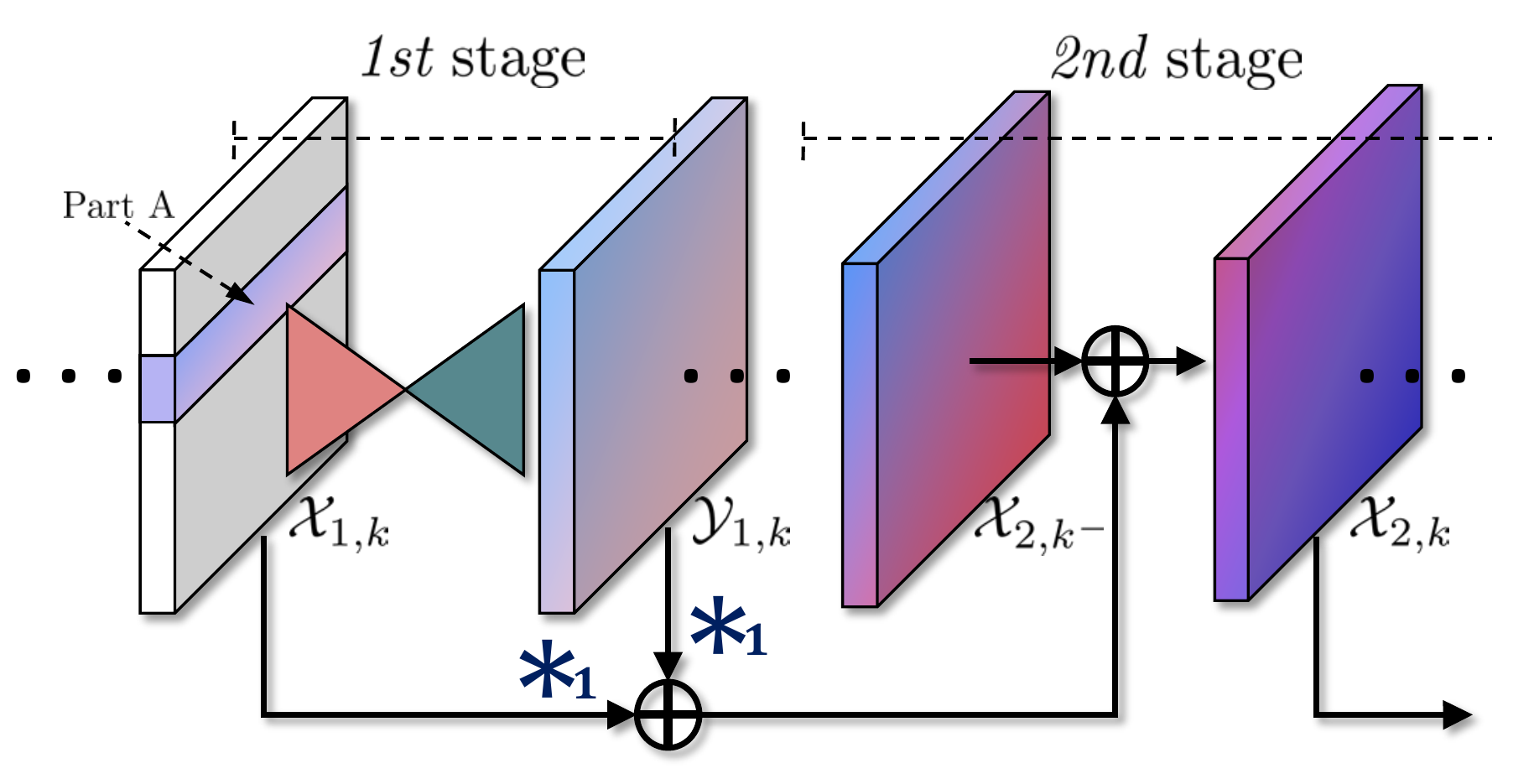}
        
    \end{subfigure}
    \caption{The procedure of the Cross Stage Feature Aggregation, especially between the first and second stages. \textcolor{black}{The colored part (Part A) denotes the region} where valid laser range information is \textcolor{black}{propagated} directly to the feature (best viewed in color).}
    \label{fig:csfa}
\end{figure}

Despite the well-balanced feature by using \textcolor{black}{the} multi-stage architecture, the architecture is associated with potential risks: it might \textcolor{black}{undergo saturation and thus become} vulnerable to \textcolor{black}{diluting} the information during repeated up and down sampling \cite{newell2016stacked}. Thus, we adopted the Cross Stage Feature Aggregation (CSFA), which is inspired by \cite{li2019rethinking, he2016deep} to alleviate this problem as follows:

\begin{equation}
\mathcal{X}_{n, k} = \mathcal{X}_{n, k^{-}} \oplus (\phi{(\mathcal{X}_{n-1, k})} \oplus \psi{(\mathcal{Y}_{n-1, k})}) 
\end{equation}
where $\phi(\cdot)$ and $\psi(\cdot)$ are learnable transformations, i.e. 1$\times$1 convolution, $\oplus$ is element-wise addition, 
$\mathcal{X}_{n, k}$ denotes the input feature of each $k$-th block of ResNet-18 in \textcolor{black}{the} $n$-th stage, and $\mathcal{Y}_{n, k}$ denotes the output feature of each $k$-th block of decoder in \textcolor{black}{the} $n$-th stage, respectively. 
$k^{-}$ is the output feature of \textcolor{black}{$(k-1)$}-th block of the encoder.

As shown in Fig. \ref{fig:csfa}, prior features from the encoder and decoder are introduced to the encoder part of the next stage. 
\textcolor{black}{By} propagating multi-scale features from early stages to the current stage, CSFA can a) prevent the network \textcolor{black}{from diluting} the feature and b) allow it to learn better \textcolor{black}{the} inter-spatial relationship \textcolor{black}{between features.} 

\subsection{Loss Function}

\textcolor{black}{To train the proposed network, different} $L_{1}$ term is used \textcolor{black}{based on the types of the input method}, i.e. \texttt{proj-d} or \texttt{ref-d}. When \texttt{proj-d} is taken as an input, the loss function is defined as follows:

\begin{equation}
\textcolor{black}{\mathcal{L}_{r} = \norm{\kappa_{\{\mathcal{D}>0\}}\cdot (\mathcal{D} -\hat{\mathcal{D}} )}_{1} }
\end{equation}  
\textcolor{black}{where $\mathcal{D}$ is the ground truth depth and $\hat{\mathcal{D}}$ is the estimated depth}. $\kappa$ denotes a pixel-wise filter that is equal to one when the pixel value of ground truth is available or zero otherwise.

\textcolor{black}{On the \textcolor{black}{contrary}, when \texttt{ref-d} is taken as \textcolor{black}{an} input, the loss function is defined as follows:}

\begin{equation}
\mathcal{L}_{r} = \norm{\kappa_{\{\mathcal{D}>0\}}\cdot (\mathcal{D} -(\hat{\mathcal{D}}_{res}+\mathcal{R}))}_{1} 
\end{equation}  
\textcolor{black}{where $\hat{\mathcal{D}}_{res}$ is the predicted residual depth from the end part of our network and $\mathcal{R}$ indicates \texttt{ref-d}.}

 Cao \textit{et al.} \cite{cao2017estimating} argued that the softmax classification loss, $\mathcal{L}_{c}$, \textcolor{black}{was} better than $\mathcal{L}_{r}$ and Liao \textit{et al.} \cite{liao2017parse} demonstrated that \textcolor{black}{the} tightly coupled loss term between $\mathcal{L}_{r}$ and $\mathcal{L}_{c}$ \textcolor{black}{was} better than $\mathcal{L}_{c}$ \textcolor{black}{only.} 
 However, we noticed that only using $\mathcal{L}_{r}$ showed better performance than \textcolor{black}{using} combination of the classification loss terms $\mathcal{L}_{c}$ and $\mathcal{L}_{r}$ from \cite{liao2017parse} in realistic environments (see Section \rom{5}.\textit{A} for its rationale).
 
\section{Experiments}

\subsection{Dataset}
\noindent 
\textbf{KAIST RGBD-scan Dataset} \quad Synchronized RGB-D and 2D LiDAR scan data were collected to evaluate accuracy and applicability of the experiment in physical indoor environments. Data were collected by a mobile robot loaded with Intel RealSense D435i \cite{intel_realsense} and Hokuyo UST-20LX \cite{hokuyo}, as shown in Fig. \ref{fig:overview_w_comparison}. Sampling frequency of the RGB-D sensor was dropped to 15Hz followed by post-processing of \textcolor{black}{the} collected data to remove overlapping scenes. Our dataset consists of 14,143 training data for 99 scenes and 1,042 test data for 53 scenes, all of which \textcolor{black}{were acquired} from KAIST campus. \textcolor{black}{Note that the distinct interior design of individual in-campus buildings makes the indoor scenes of KAIST much appropriate for a dataset.}

\noindent 
\textbf{KITTI Odometry Dataset} \quad \textcolor{black}{Unfortunately, RealSense D435i depth sensor is too noisy to be qualified as a ground truth for evaluating depth prediction \cite{d435_fluc}.} Thus, for clear comparison, our network was also trained and evaluated \textcolor{black}{using} KITTI \cite{geiger2012we}, which is an outdoor dataset. We follow\textcolor{black}{ed} \textcolor{black}{the procedure} in \cite{mal2018sparse} and simulated 2D laser scans \textcolor{black}{were} sub-sampled \textcolor{black}{from} the Velodyne 64E 3D scanner based on \cite{liao2017parse}.
 
\noindent 
\textbf{NCLT Dataset} \quad The University of Michigan North Campus Long-Term Vision and LiDAR (NCLT) dataset provides synchronized 3D LiDAR scans, 2D LiDAR scans, and images \textcolor{black}{in} both indoor and outdoor environments \cite{carlevaris2016university}. \textcolor{black}{This dataset was} utilized to check which input method type, i.e. \texttt{ref-d} or \texttt{proj-d}, is more robust \textcolor{black}{under} untrained scenarios. 9,733 samples from three categories (``2012-01-08'', ``2012-03-25'', ``2012-05-11'') \textcolor{black}{are} used for \textcolor{black}{the} training dataset and 1,050 samples from the four categories (``2012-01-15'', ``2012-03-31'', ``2012-08-04'', ``2012-11-04'') were used for \textcolor{black}{the} test dataset. \textcolor{black}{Note that all training data were captured only in daytime, \textcolor{black}{e.g.} sunny or partly cloudy weathers, whereas the test data additionally includes} \textcolor{black}{some scenes with foliage and some in snowy and cloudy weathers.}
\textcolor{black}{Because of} the relative\textcolor{black}{ly} small training set, the network was initialized by the weights learned from KITTI.

\subsection{Training the Network}

To train our network, the Adam optimizer\cite{kingma2014adam} was exploited for 30 epochs with learning rate of 0.0001, momentum of 0.9, weight decay of 0.0001, decay rate of 0.98, and batch size of 20. Note that weight decays for every epoch. Our network was modeled by PyTorch and \textcolor{black}{was} trained with two NVidia TITAN Xp GPUs.

\subsection{Error Metrics}

For evaluation, we followed some metrics commonly utilized in depth prediction area \cite{liao2017parse, mal2018sparse, cheng2018depth}. Let $\mathcal{D}_l$ and $\hat{\mathcal{D}_l}$ be the ground truth and \textcolor{black}{the} estimated depth on each pixel $l$, respectively. Then metrics are as following:

\begin{itemize}
    \item Root Mean Squared Error (RMSE): $\sqrt{\frac{1}{N}\sum_{l}(\hat{\mathcal{D}_l}-\mathcal{D}_l)^2}$
    
    \item Mean Absolute Relative Error (REL): $\frac{1}{N}\sum_{l}\frac{|{\hat{\mathcal{D}_l}-\mathcal{D}_l}|}{\mathcal{D}_l}$
    
    \item $\delta_{n}$: percentage of $\mathcal{D}_l$, such that max($\frac{\hat{\mathcal{D}_l}}{\mathcal{D}_l}$, $\frac{\mathcal{D}_l}{\hat{\mathcal{D}_l}}$) < $\delta^n$, $\delta$ = 1.25 and $n$ = 1, 2, 3.
    
\end{itemize}
 where $N$ is the total number of pixels. 

\section{Results and Discussion}
 
 Keep in mind that \textcolor{black}{the visualization of all the sparse depth measured from both 2D and 3D LiDAR sensors has been magnified} for \textcolor{black}{ease of} understanding. Originally, each range data \textcolor{black}{is} represented in only one pixel. \textcolor{black}{As} previous works \cite{liao2017parse, lee2019depth, mal2018sparse} \textcolor{black}{show} that \texttt{RGB} with some sparse depth leads to better depth prediction results compared to those from \textcolor{black}{\texttt{RGB} only, we instead gave more emphasis on the comparison of encoding methods \textcolor{black}{for} 2D LiDAR scans, i.e. \texttt{ref-d} and \texttt{proj-d}. Thus, the symbol \texttt{RGB} is omitted in front of the symbols \texttt{ref-d} and \texttt{proj-d} for simplicity in our discussion.}

\subsection{Ablation Study}

\noindent 
\textbf{Loss Function} Through the ablation study, it is noted that combination of the loss terms with \texttt{ref-d} shows \textcolor{black}{more} promising performance \textcolor{black}{than that with \texttt{proj-d}}. However, for our proposed networks, it is shown that only using $\mathcal{L}_{r}$ leads to better performance than using combination of $\mathcal{L}_{r}$ and $\mathcal{L}_{c}$ as shown in Table \ref{table:ablation_loss}. 

\begin{table}[h]
    \centering
    \caption{Ablation study: performance with different loss functions and input method types on the KAIST RGBD-scan dataset when the number of stage $N$=2.}
    \begin{tabular}{ccccccc}
		\toprule
	Loss Function   &   Input method &  RMSE [mm]  &      REL & $\delta_{1}$ [\%] \\ \midrule
	\multirow{2}{*}	{$L_{r} + L_{c}$}     
	 & \texttt{proj-d}  &  540   &     0.122    &  88.4  \\
     & \texttt{ref-d}   &  504   &     0.101   &   89.9  \\ \midrule
      \multirow{2}{*}	{$L_{r}$ only} 
      & \texttt{proj-d}  &    496     &   0.098    &      90.0  \\
      & \texttt{ref-d}   &    \textbf{491}   &  \textbf{0.096}   &   \textbf{90.5}  \\
       \bottomrule

	\end{tabular}
	
	\label{table:ablation_loss}
\end{table}

\noindent 
\textbf{Cross Stage Feature Aggregation} \quad After choosing the loss function as $\mathcal{L}_{r}$, \textcolor{black}{the effect of CSFA on the performance is analyzed.} 
As depicted in \textcolor{black}{Table \ref{table:csfa}}, it is shown that CSFA lead\textcolor{black}{s} to significant performance improvement.

It is shown that \textcolor{black}{na\"ively stacking the two networks ($N$=2) demonstrates rather worse performance than having one network ($N$=1). Besides, employing 1$\times$1 convolutions \textcolor{black}{enhances} the depth accuracy and leads to larger performance improvement than merely connecting the networks.} Thus, the result impl\textcolor{black}{ies} that CSFA mitigate\textcolor{black}{s} the saturation issue of multi-stage networks in depth prediction tasks.

\begin{table}[h]
    \centering
    \caption{Ablation study: performance \textcolor{black}{comparison based on the presence or absence of each component of CSFA} on the KAIST RGBD-scan dataset.}
    \begin{tabular}{c|ccccc}
		\toprule
	Stages & Connection     & Conv.   &  RMSE [mm]  &   REL   & $\delta_{1}$ [\%] \\ \midrule
	 1$\times$Res-18 &              &        &  512          &       0.107      &  89.2  \\  \midrule
	 \multirow{3}{*}{2$\times$Res-18} &      &          &  516          &       0.107      &  89.2  \\  
	       & \checkmark  &          &   506  &       0.103      &  89.9   \\  
           &  \checkmark &\checkmark&            491 &  \textbf{0.096}   &   \textbf{90.5}  \\ \midrule
     4$\times$Res-18 &  \checkmark &\checkmark&  \textbf{485} &  0.098   &   90.0  \\      \bottomrule

	\end{tabular}
	\label{table:csfa}
\end{table}

\noindent 
\textbf{Multi-stage Architecture} \quad As reported in \textcolor{black}{Table \ref{table:csfa}}, the increasing number of $N$ \textcolor{black}{leads} to performance improvement. Specifically, our 2-stage network led to 21mm improvement over 1-stage network and obtained 491mm of RMSE. These experiments indicate that multi-stage architecture successfully mitigates the partial observation issues for better depth prediction in terms of \textcolor{black}{errors.} Additionally, estimated depth from \textcolor{black}{the 4-stage network has} smaller errors with \textcolor{black}{the reliable} geometry compared to 2-stage network, yet its performance gain \textcolor{black}{is} decreased. 

\begin{figure}[h]
	\centering
	\begin{subfigure}[b]{0.23\textwidth}
		\includegraphics[width=1.0\textwidth]{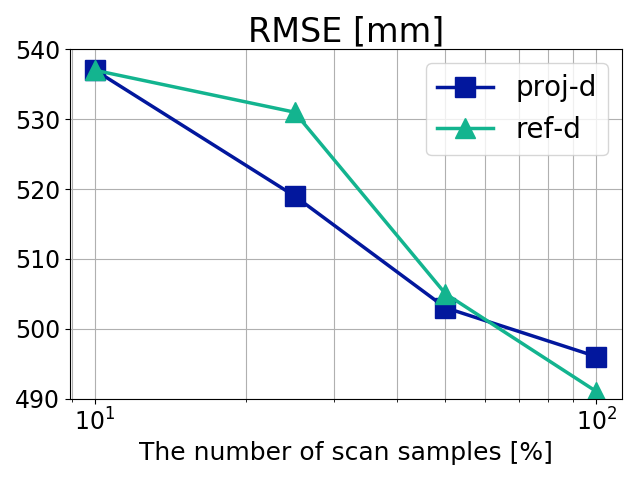}
		\caption{}
		\label{fig:num_of_scans_rmse}
	\end{subfigure}
	\begin{subfigure}[b]{0.23\textwidth}
		\includegraphics[width=1.0\textwidth]{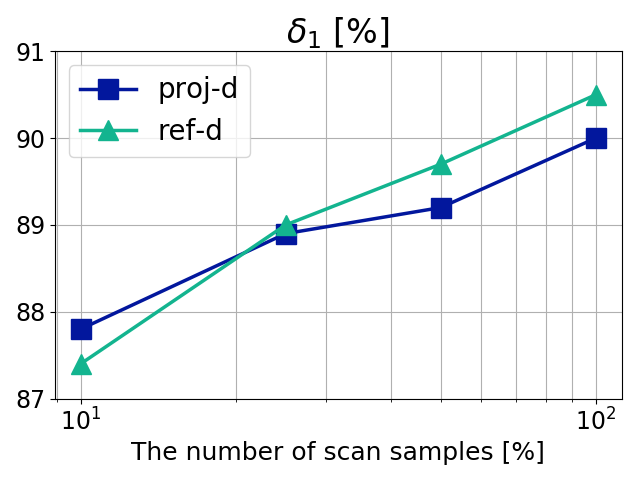}
		\caption{}
		\label{fig:num_of_scans_delta}
	\end{subfigure}
	
	\caption{Performance changes with \textcolor{black}{varying} number of scan samples. (a) RMSE (\textit{the lower, the better}) (b) $\delta_{1}$ (\textit{the higher, the better}) with the scans dropout.}
	\label{fig:num_of_scans}
\end{figure}

\noindent 
\textbf{Impact of the number of 2D Laser Scans} \quad As reported in Table \textcolor{black}{\ref{table:kaist_comparison}}, taking  \texttt{RGB} with \texttt{ref-d} as input led to smaller error compared to \texttt{RGB} with \texttt{proj-d}. Accordingly, \textcolor{black}{performance analysis} with \textcolor{black}{different} number of laser scans are conducted on KAIST RGBD-scan dataset. As described in Fig. \ref{fig:num_of_scans}, the \texttt{ref-d} outperforms \texttt{proj-d} with over 50\% of scan dropout.
On the other hand, \texttt{proj-d} \textcolor{black}{has} a tendency to outperform \texttt{ref-d} as \textcolor{black}{the} sample size decreases under 50\%.

Thus, we could draw a conclusion that one should choose the encoding method \textcolor{black}{depending} on one's circumstances:
a) \texttt{ref-d} is suitable for 2D LiDARs with high resolution, e.g. angular resolution is \textcolor{black}{higher than} $\ang{0.25}$ like Hokuyo UST-20LX \cite{hokuyo}.
b) \texttt{proj-d} is suitable for those with less dense resolution, e.g. angular resolution is \textcolor{black}{lower than} $\ang{1.0}$ like LDS-01 sensor\cite{ldslidar} in realistic environments.

\subsection{Robustness on the \textcolor{black}{Untrained} Scenarios }
\begin{figure}[h]
    \centering
    \begin{tabular}[t]{cc}
        \centering
        \begin{subfigure}[b]{.5\linewidth}
            \includegraphics[width=\linewidth]{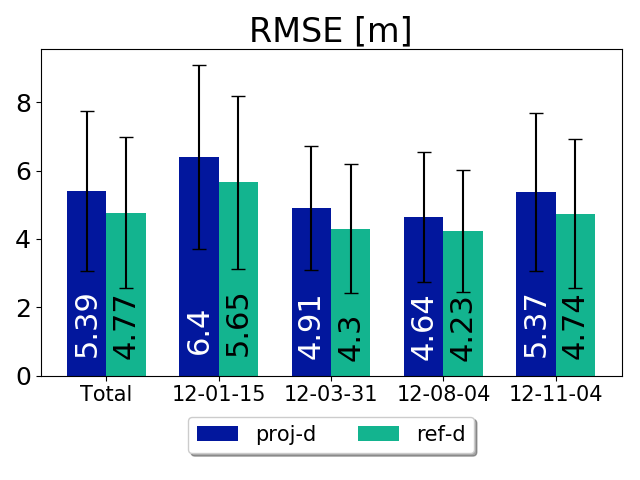}
            \caption{}\label{fig:robust_rmse}
        \end{subfigure}
        
        \begin{subfigure}[b]{.5\linewidth}
            \includegraphics[width=\linewidth]{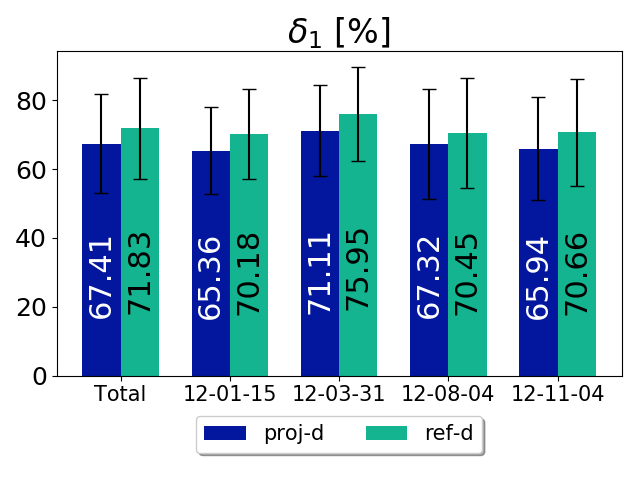}
            \caption{}\label{fig:robust_delta}
        \end{subfigure}
        
    \end{tabular}
    \caption{\textcolor{black}{Performances with respect to different input methods.} (a) RMSE (\textit{the lower, the better}) (b) $\delta_{1}$ (\textit{the higher, the better}) \textcolor{black}{with each date class of the test data on NCLT dataset.} }
    \label{fig:robust_errors}
\end{figure}

In \textcolor{black}{untrained} scenarios, which are not included in the train dataset, accuracy of \texttt{ref-d} is better \textcolor{black}{on} every date as described in Fig. \ref{fig:robust_errors}. Specifically, the \texttt{ref-d} outperforms \texttt{proj-d} with 4.77m of RMSE and 71.83\% of \textcolor{black}{$\delta_{1}$.} Especially, the performance gap also increases on the ``12-01-15'' test data, which are captured on snowy day so that there exists huge photometric differences \textcolor{black}{with the training dataset. }

\begin{figure*}[h]
	\centering
	
	\begin{subfigure}[b]{.95\linewidth}
        \includegraphics[width=\linewidth]{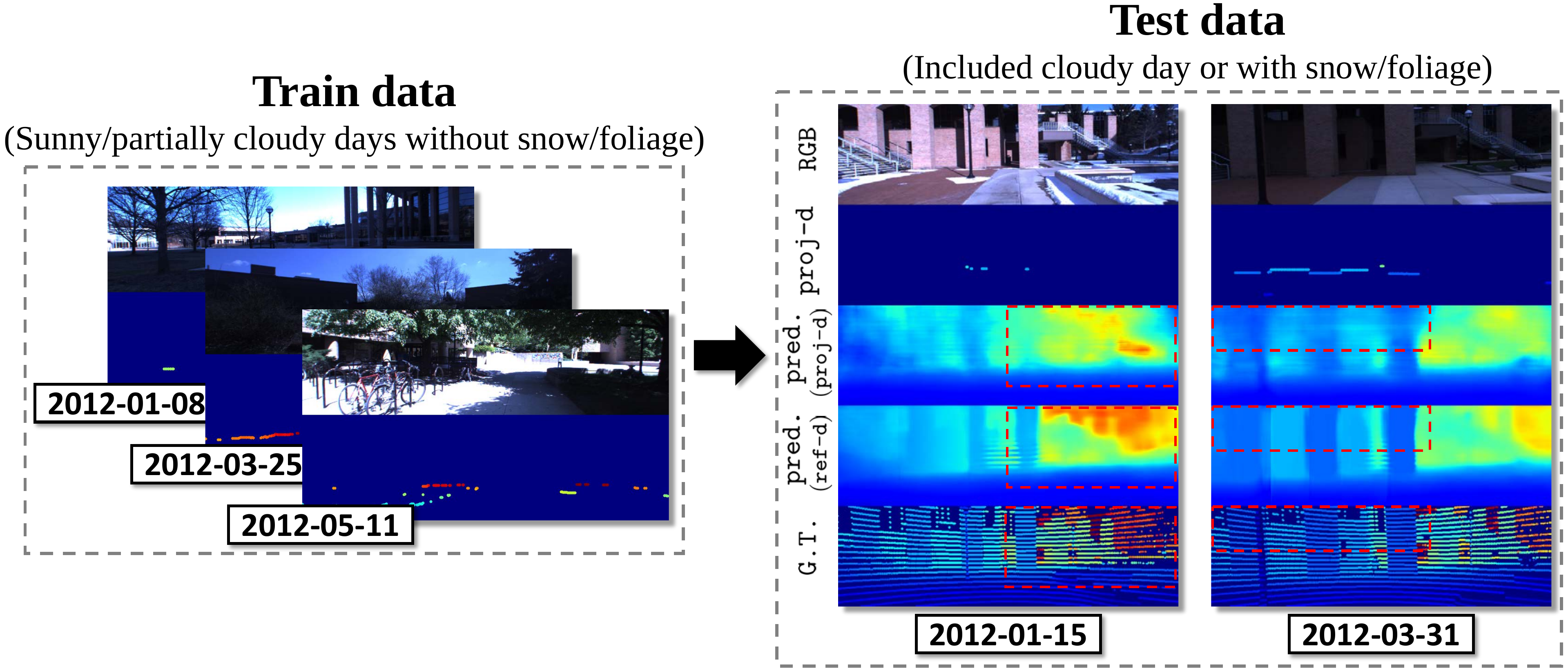}
    \end{subfigure}
    
	\caption{\textcolor{black}{An overview of robustness test with different types of input method and visualized results on NCLT dataset.} }\label{fig:nclt_whole_system}
\end{figure*}

  The visualized results are presented in Fig. \ref{fig:nclt_whole_system}. The tested scenes are also included in the train dataset with different weathers so that it becomes hard to predict depth precisely.
  \textcolor{black}{Both results of \texttt{ref-d} and \texttt{proj-d} imply that they both can precisely estimate the depth of a region where the laser actually scans. However, since \texttt{ref-d} propagates geometrical information gathered by the laser scan vertically, the network with  \texttt{ref-d} can estimate more refined depth of an image that has objects with vertically identical depth. Thus, the experiment shows that taking \texttt{ref-d} is more robust under \textcolor{black}{untrained} conditions, especially for realistic environments where geometrically structured objects are present.}

\subsection{Comparison with State-of-the-arts}

 \textcolor{black}{O}ur best model investigated in Section \rom{5}.\textit{A} \textcolor{black}{was compared} with the CNN-based state-of-the-art methods quantitatively, Liao \textit{et al.} \cite{liao2017parse}, Ma and Karaman \cite{mal2018sparse}, and Cheng \textit{et al.}\cite{cheng2018depth}, trained on the KAIST RGBD-scan dataset and KITTI.

\begin{figure*}[h]
    \centering
    \begin{tabular}[t]{cccccc}
        \centering
        \begin{subfigure}[b]{.15\linewidth}
            \includegraphics[width=\linewidth]{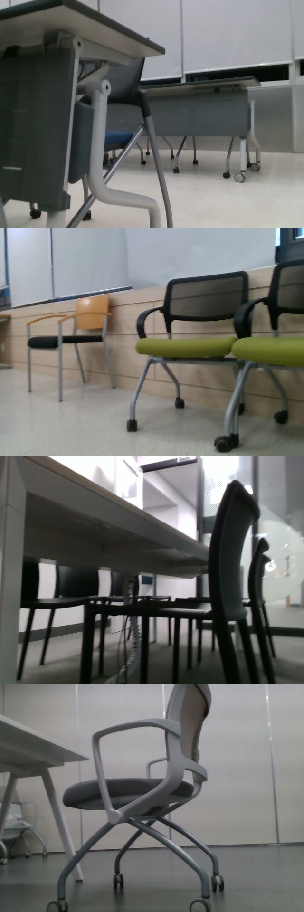}
            \caption{RGB}\label{fig:comp_rgb}
        \end{subfigure}
        
        \begin{subfigure}[b]{.15\linewidth}
            \includegraphics[width=\linewidth]{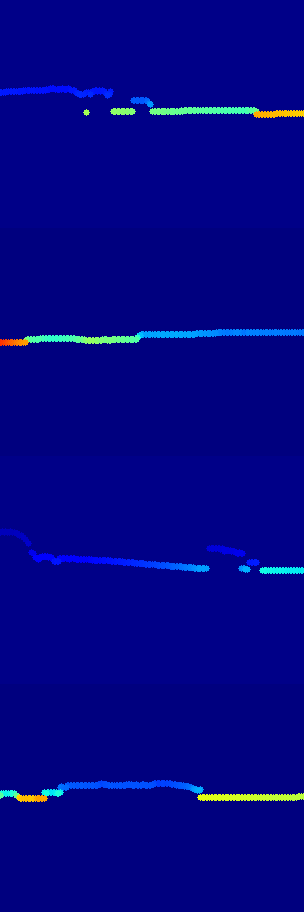}
            \caption{Laser scans}\label{fig:comp_scan}
        \end{subfigure}
        
        \begin{subfigure}[b]{.15\linewidth}
            \includegraphics[width=\linewidth]{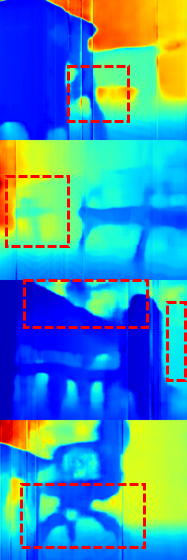}
            \caption{\cite{liao2017parse}}\label{fig:comp_gt}
        \end{subfigure}
        
        \begin{subfigure}[b]{.15\linewidth}
            \includegraphics[width=\linewidth]{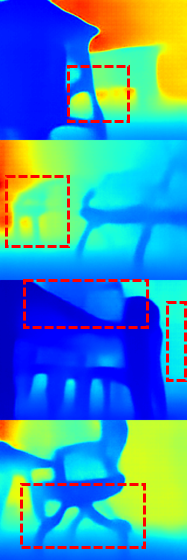}
            \caption{\cite{mal2018sparse}}\label{fig:comp_liao}
        \end{subfigure}
        
        \begin{subfigure}[b]{.15\linewidth}
            \includegraphics[width=\linewidth]{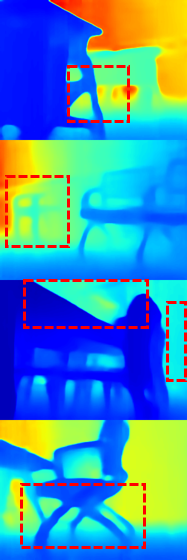}
            \caption{Ours}\label{fig:comp_ma}
        \end{subfigure}
        
        \begin{subfigure}[b]{.15\linewidth}
            \includegraphics[width=\linewidth]{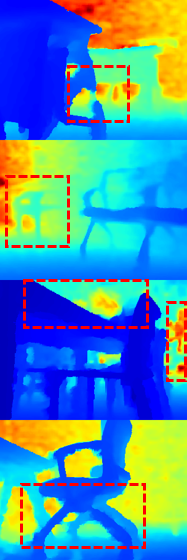}
            \caption{Ground truth}\label{fig:comp_ours}
        \end{subfigure}

    \end{tabular}
    \caption{Comparison of depth \textcolor{black}{estimation results with state-of-the-arts on KAIST RGBD-scan dataset.} }
    \label{fig:comp_kaist}
\end{figure*}

\noindent 
\textbf{KAIST RGBD-Scan Dataset} \quad The results on KAIST RGBD-scan dataset are reported in Table \ref{table:kaist_comparison} and shown in Fig. \ref{fig:comp_kaist}. We can see that results of \cite{liao2017parse} and \cite{mal2018sparse} give acceptable and precise depth maps, \textcolor{black}{filtering out undesirable irradiation errors caused by the characteristics of the RGB-D camera}. As reported in Table \ref{table:kaist_comparison}, our MSDPN exhibits \textcolor{black}{most} promising results with more sharp edges and refined depths \textcolor{black}{for both types of input}. Specifically, our method yields smaller errors with \textcolor{black}{fewer} parameters and FLOPs.

\begin{table*}[h]
	\centering
	\caption{Comparison with state-of-the-arts on the KAIST RGBD-scan dataset}
	\begin{tabular}{lllccccccc}
		\toprule
		\multirow{2}{*}{Input} & \multirow{2}{*}{Method}&\multirow{2}{*}{Backbone} &\multirow{2}{*}{Params} &\multirow{2}{*}{FLOPs}  & \multicolumn{2}{c}{Error \textit{(Lower the Better)}} & \multicolumn{3}{c}{Accuracy \textit{(Highter the better)}} \\ \cmidrule(lr){6-7} \cmidrule(lr){8-10}
		
                        & & & & \multicolumn{1}{c}{}& RMSE [mm]   & REL         & $\delta_{1}$ [\%]  & $\delta_{2}$ [\%]  & $\delta_{3}$ [\%] \\ \midrule
	
		\multirow{3}{*}{\texttt{RGB+proj-d}}&\multirow{1}{*}{Liao \textit{et al.} \cite{liao2017parse}} & Res-50&\multirow{1}{*}{48.7M}  & \multirow{1}{*}{39.8G} &  640   & 0.132  & 83.8  & 93.7  & 97.0 \\ 
		
		&\multirow{1}{*}{Ma and Karaman \cite{mal2018sparse}} & Res-50&\multirow{1}{*}{63.6M}  & 
		\multirow{1}{*}{42.8G} & 504  & 0.102 & 89.7 & 96.0 & 98.0 \\ 
		
		& MSDPN (Ours) &2$\times$Res-18& \textbf{47.4M}  & \textbf{30.0G} &   \textbf{496}   &   \textbf{0.098}    &  \textbf{90.0}  & \textbf{96.1}  & \textbf{98.0} \\ \midrule
		
		\multirow{3}{*}{\texttt{RGB+ref-d}}&\multirow{1}{*}{Liao \textit{et al.} \cite{liao2017parse}} & Res-50&\multirow{1}{*}{48.7M}  & \multirow{1}{*}{39.8G} &  534   & 0.109  & 87.9  & 95.4  & 97.8 \\ 
		
		&\multirow{1}{*}{Ma and Karaman \cite{mal2018sparse}} & Res-50&\multirow{1}{*}{63.6M}  & 
		\multirow{1}{*}{42.8G} & 516  & 0.107 & 89.2 & 95.8 & 98.0 \\ 
		
		& MSDPN (Ours) &2$\times$Res-18& \textbf{47.4M}  & \textbf{30.0G} &   \textbf{491}   &  \textbf{0.096}  &  \textbf{90.5}  & \textbf{96.3}  & \textbf{98.2} \\ \bottomrule
	\end{tabular}
	\label{table:kaist_comparison}
\end{table*}

Furthermore, it was checked that the MSDPN propagates the depth information effectively to upward and downward regions so that \textcolor{black}{it} learns inter-spatial relationships better \textcolor{black}{than} other state-of-the-arts methods.
As shown in \textcolor{black}{the} upper bounding box of the third row of Fig. \ref{fig:comp_kaist}, \textcolor{black}{Due to its own propagation method and multi-stage architecture, our MSPDN estimates more precise depth of the region where direct depth cues are missing than other single-stage networks}. Therefore, it is verified that repeated upsampling and downsampling of feature in multiple encoder-decoder architecture mitigates the partial observation issue.

\begin{figure}[h]
    \centering
    \begin{subfigure}[b]{.95\linewidth}
        \includegraphics[width=\linewidth]{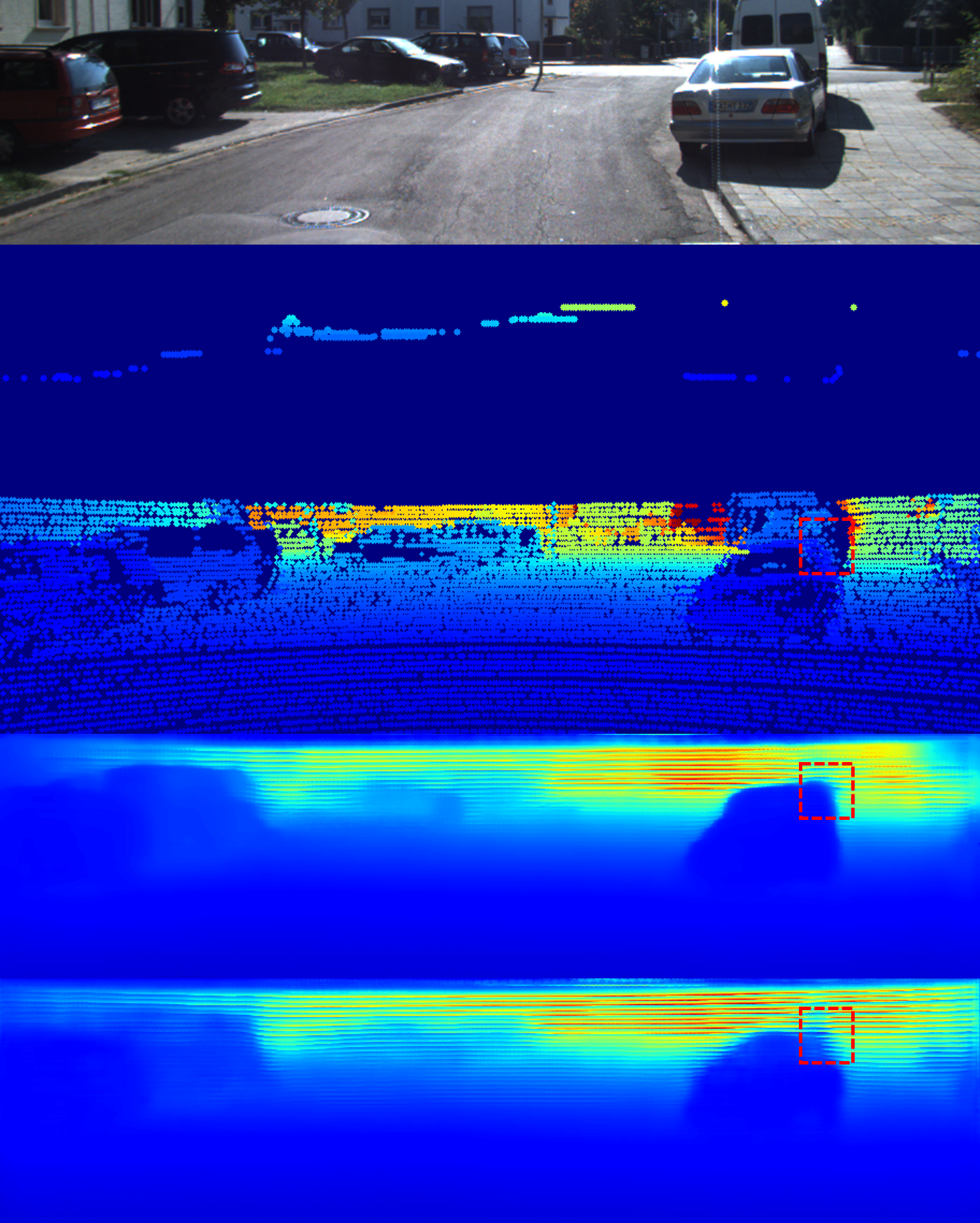}
        
    \end{subfigure}
    \caption{Qualitative comparison with state-of-the-arts on KITTI (T-B): \texttt{RGB}, simulated 2D laser scans (\texttt{proj-d}), ground truth, Cheng \textit{et al.} \cite{cheng2018depth}, and our (best viewed in color).}
    \label{fig:kitti}
\end{figure}

\noindent 
\textbf{KITTI Odometry Dataset} \quad All quantitative results \textcolor{black}{for KITTI dataset} are reported in Table \ref{table:kitti_comparison}. We trained \cite{mal2018sparse}, \cite{cheng2018depth}, and ours on KITTI. 
As shown in Fig. \ref{fig:kitti}, the experiments showed that all state-of-the-arts alleviate the global ambiguity issue by utilizing simulated 2D LiDAR scans compared to those that take \texttt{RGB} as input. 
Among them, our MSDPN yields smaller RMSE and higher $\delta_{1}$ compared to other methods. 
Furthermore, it is noted that our MSDPN presents significant depth accuracy when taking \texttt{RGB} as input compared to other methods. Therefore, we draw a conclusion that MSDPN not only somehow address\textcolor{black}{es} the partial distribution issue effectively, but also \textcolor{black}{proves itself to be an effective way} to predict depth in general.

\begin{table}[h]
    \centering
    \caption{Comparison with state-of-the-arts on the KITTI Odometry dataset. Results of \texttt{RGB} are quoted from \cite{mal2018sparse}.}
	\begin{tabular}{llccccc}
		\toprule
		Input   &   Method &  RMSE [m]  &      REL & $\delta_{1}$ [\%] \\ \midrule
	\multirow{5}{*}	{\texttt{RGB}}     
	 & Make3D \cite{laina2016deeper}   &  8.734     &     0.280    &   60.1  \\
     & Mancini \cite{mancini2016fast}  &  7.508     &     -    &        31.8  \\
     & Eigen \textit{et al.} \cite{eigen2014depth}  &    7.156 &    0.190  & 69.2  \\
     & Ma and Karaman \cite{mal2018sparse}   &  6.266   &  0.208  &     59.1  \\
     & MSDPN (Ours) &      \textbf{4.656}     &    \textbf{0.104}   &   \textbf{87.3}  \\ \midrule
      \multirow{5}{*}	{\texttt{RGBd}} 
      & Liao \textit{et al.} \cite{liao2017parse}  &    4.500    &   0.113    &     87.4  \\
      & Ma and Karaman \cite{mal2018sparse}        &    4.093    & \textbf{0.084}    &  91.1  \\
      & Cheng \textit{et al.} \cite{cheng2018depth} &    4.017   &  0.090   &   90.9  \\
      & MSDPN (Ours)       &    \textbf{3.834}   &  0.090   &   \textbf{91.5}  \\
       \bottomrule

	\end{tabular}
	\label{table:kitti_comparison}
\end{table}

\section{Conclusion}

In this study, an encoder-decoder-based multi-stage deep learning architecture, \textcolor{black}{namely} Multi-\textcolor{black}{S}tage Depth Prediction Network (MSDPN)\textcolor{black}{,}  has been proposed for efficient depth prediction from 2D LiDAR scans and a monocular image. We also \textcolor{black}{suggested} Cross Stage Feature Aggregation to prevent the network \textcolor{black}{from saturating and to allow} the network to learn inter-spatial relationships within the features. 
Our proposed network \textcolor{black}{was tested} on realistic environments: using not only \textcolor{black}{the} sub-sampled input data but also \textcolor{black}{physically collected dataset} from \textcolor{black}{a} 2D LiDAR. 
Accordingly, through the ablation study and \textcolor{black}{the} comparison with state-of-the-art methods, the effectiveness of the MSDPN was verified and it was shown that our proposed method yielded the most precise depth estimations both quantitatively and qualitatively. 
Finally, the robustness of the \textcolor{black}{different} input methods are also tested \textcolor{black}{in} the untrained situations.

In future works, our loss function will be investigated to refine \textcolor{black}{depth} better, by introducing consistencies as suggested in \cite{lee2019depth}. Furthermore, this approach could be applied \textcolor{black}{to} path planning of mobile robot platforms. Additionally, since the mechanism of our system is similar to the structured light range finder systems\cite{jeon2017one, fisher1999low}, we will check whether our system is applicable to the structured light systems or not.

\bibliographystyle{IEEEtran}
\bibliography{./iros20,./IEEEabrv}

\end{document}